\documentclass[runningheads]{llncs}
\usepackage[T1]{fontenc}
\usepackage{multirow}
\usepackage{graphicx}
\usepackage{booktabs}
\usepackage{verbatim}
\usepackage{amsmath}
\begin{document}
\title{Detecting LGBTQ+ Instances of Cyberbullying}

\author{ Muhammad Arslan \inst{1} \and
Manuel Sandoval Madrigal \inst{1} \and
Mohammed Abuhamad \inst{1} \and
Deborah L. Hall \inst{2} \and
Yasin N. Silva \inst{1}}
\authorrunning{M. Arslan et al.}
\institute{Loyola University Chicago, Chicago IL 60626, USA 
\email{\{marslan,msandovalmadrigal,mabuhamad,ysilva1\}@luc.edu} \and
Arizona State University, Glendale, AZ 85306 \\
\email{d.hall@asu.edu}}

%\author{First Author\inst{1}\orcidID{0000-1111-2222-3333} \and
%Second Author\inst{2,3}\orcidID{1111-2222-3333-4444} \and
%Third Author\inst{3}\orcidID{2222--3333-4444-5555}}
%\institute{Princeton University, Princeton NJ 08544, USA \and
%Springer Heidelberg, Tiergartenstr. 17, 69121 Heidelberg, Germany

\maketitle           
\begin{abstract}
Social media continues to have an impact on the trajectory of humanity. However, its introduction has also weaponized keyboards, allowing the abusive language normally reserved for in-person bullying to jump onto the screen, i.e., cyberbullying. Cyberbullying poses a significant threat to adolescents globally, affecting the mental health and well-being of many. A group that is particularly at risk is the LGBTQ+ community, as researchers have uncovered a strong correlation between identifying as LGBTQ+ and suffering from greater online harassment. Therefore, it is critical to develop machine learning models that can accurately discern cyberbullying incidents as they happen to LGBTQ+ members. The aim of this study is to compare the efficacy of several transformer models in identifying cyberbullying targeting LGBTQ+ individuals. We seek to determine the relative merits and demerits of these existing methods in addressing complex and subtle kinds of cyberbullying by assessing their effectiveness with real social media data.

\end{abstract}

\section{Introduction}
Cyberbullying continues to be a pressing issue that affects a considerable proportion of adolescents across the world. It is estimated that nearly 10\% to 20\% of adolescences have experienced cyberbullying at some time, and if left unaddressed, cyberbullying can contribute to significantly increased risk for suicide among adolescents \cite{Ginanluca_2014}. That is, while the exact relationship between bullying and suicidality (suicidal ideation or suicide attempts) is complex, there is evidence to suggest that exposure to peer bullying and peer aggression are risk factors \cite{bullying_suicidal_ideation_meta_analysis_2015}. As with many of the technological advancements of our day, bullying has also evolved into cyberbullying, the use of technology/electronic media as a means to bully or harass. One way to address cyberbullying is through the development of machine learning-based moderation tools that readily scale for the task of cyberbullying detection. While prior research in cyberbullying detection has focused primarily on the issue of general cyberbullying, limited work has been done towards developing LGBTQ+ sensitive models. Specifically, most cyberbullying detection models approach cyberbullying as one-size-fits-all, while the reality is that the type of harassment experienced by LGBTQ+ members varies significantly from that of the general public. Even more concerning is that sexual and gender identity minorities are significantly more likely to be victims of cyberbullying and online harassment compared to heterosexual and cisgender individuals \cite{bully_cyberbullying_lgbtq_pdf}. This strife is then compounded as LGBTQ+ members are more severely impacted by these negative interactions due to well-documented disparities in mental health and social support and the chronic experiences of minority stress \cite{ploderl_tremblay_2015}. With cyberbullying corresponding with an increase in suicidality amongst the general public, we see even more pressing need for cyberbullying models that support the LGBTQ+ community. \\

In this paper, we describe the implementation and performance evaluation of several large language-based models that aim to identify cyberbullying posts that target LGBTQ+ individuals. We seek to determine the relative merits and demerits of these existing methods in addressing complex and subtle kinds of cyberbullying by assessing their effectiveness with an Instagram dataset.

\section{Related Work}

\textbf{Cyberbullying Detection via Machine Learning.} The domain of general cyberbullying detection has received considerable empirical attention, in terms of efforts to adapt the latest methods from the machine learning literature and to propose models that take advantage of features specific to each social media platform. Early contributions to cyberbullying detection focused on applying off-the-shelf solutions, e.g., SVM, Naïve Bayes, and Logistic Regression, to standard binary classification (bullying versus non-bullying) \cite{MDavar2012}. Well-established deep learning architectures, such as CNN, LSTM, BiLSTM, and BiLSTM with attention, were studied and applied by Dadvar and Eckert to a cyberbullying-labeled YouTube dataset that included 53k posts and 4k users \cite{dadvar2018cyberbullying}. Cheng et al. proposed the inclusion of network-related content such as user profile information, likes, and follows to identify cyberbullying \cite{cheng_pi-bully_2019}. In other studies by Cheng et al. \cite{cheng_hierarchical_2019,lu_cheng_2021_hant}, they studied how the temporal dynamics of comment arrival can be modeled as part of a hierarchical attention network. Some researchers have also proposed multi-modal models that incorporate video, images, and time-related components into the architectures of their models \cite{singh_toward_2017}.

\noindent
\textbf{Cyberbullying Risk among LGBTQ+ Users.}
A robust finding within social science research is the disproportionately high rates at which LGBTQ+ individuals are targets of bullying, harrassment, and discrimination---including cyberbullying (e.g., \cite{bully_cyberbullying_lgbtq_pdf}). For example, in a national sample of U.S. teens, more than half of those identifying as LGBTQ+ (52\%) %can someone help convert the word percent to the symbol
had been a victim of cyberbullying compared to 35\% of Non-LGBTQ+ teens \cite{bully_cyberbullying_lgbtq_pdf}. The importance of efforts to help identify and quickly respond to cyberbullying targeting LGBTQ+ users is underscored by mental health disparities that are distinct from and also exacerbated by cyberbullying experiences. That is, relative to their Non-LGBTQ+ peers, LGBTQ+ individuals experience poorer mental health outcomes \cite{ploderl_tremblay_2015}, due in part to chronic experiences of minority stress and broader socioeconomic and health disparities (\cite{Durrbaum_2020}), 
and are more likely to suffer negative mental health consequences, including depression, as a result of cyberbullying \cite{Duarte2018}. 

\noindent
\textbf{Toward LGBTQ+ Cyberbullying Detection.}
Most of the previous work on cyberbullying detection has focused on the development of generic models. However, recent work highlighted potential biases against certain users that can be introduced by available datasets and previous models \cite{cheng-etal-2021-mitigating}. For instance, this work reported that 68.4\% of sessions containing the word “gay” were labeled bullying in a commonly used Instagram dataset \cite{hosseinmardi2015detection}. Moreover, Tangila et al. highlighted the importance of tailored online safety tools after evaluating an Instagram dataset and finding that LGBTQ+ teens experienced significantly more high-risk online interactions than their heterosexual peers \cite{Tanni_2024}.
More recently, some initial work aimed at automatically detecting homophobia and transphobia in YouTube comments \cite{chakravarthi2021dataset} reported the results of a shared task conducted as part of a workshop in homophobia and transphobia detection (LTEDI-ACL 2022). The top-2 results for English obtained F1 Macro scores were 57\% and 49\% (based on the work by Maimaitituoheti et al. \cite{maimaitituoheti-2022-ablimet} and Sammaan et al. \cite{Tanni_2024}, respectively), highlighting the need for additional work in this area.

\section{Methods}

\noindent
\textbf{Problem Definition.}
Let \(C = \{p_1, p_2, \ldots, p_n\}\) be a corpus of \(n \) samples, where \(p_i = \{w_1, w_2, \ldots, w_l\}\) is the \( l \)-length tokenized representation of a given comment. Let \(Y = \{y_1, y_2, \ldots, y_n\}\) be the associated labels for samples in \( C \) where each $p_i \rightarrow y_i$. For any label \(y_i \in \{0, 1\}\), we have $y_i = 1$ when a comment contains LGBTQ+ related cyberbullying and $y_i = 0$ when a comment does not. The goal is to train a LGBTQ+-sensitive classifier $f$ such that \( f(p) \rightarrow y \).

\begin{table}[h]
    \centering
    \begin{tabular}{|p{0.45\textwidth}|p{0.45\textwidth}|}
        \hline
        \textbf{LGBTQ+ Related Comments} & \textbf{Non-LGBTQ+ Related Comments} \\
        \hline
        Hey everyone this faggot @\textbf{username1} is following 666 people. & @\textbf{username2} lmao just because ur an elounor shipper doesn't mean you have to be a bitch lol shut up \\
        \hline
        He must be illuminati right? Do your research before you input your opinion. You sound ignorant. & And one more thing @\textbf{username3} whoooooo cares about your st**** boyfriend.....whell i think NONE OF USS exept u :) \\
        \hline
        Ur gay dawg like really & Fuck you! And btw, it was the X factor nimrod. She can sing better than you will in your life time. So stfu. @\textbf{username4} \\
        \hline
        PUNK ASS DADDY FAG & Ur a white piece of trash \\
        \hline
        Gay people disgust me and I hate them. They scared me! That's all I'm gunna say! &  Shove off baby ugly @\textbf{username5} \\
        \hline
    \end{tabular}
    \caption{A sample of both LGBTQ+ and non-LGBTQ+ cyberbullying comments. Usernames were anonymized.}
    \label{tab:dataset_example}
\end{table}

\noindent
\textbf{Dataset.}
The dataset used in this research was obtained from the study "A Labeled Dataset for Investigating Cyberbullying Content Patterns in Instagram" \cite{Hamlett_Powell_Silva_Hall_2022}. The dataset contains 1,083 Instagram comments related to cyberbullying, with 217 comments specifically targeting the LGBTQ+ community. The dataset was annotated by a diverse team of annotators with expertise in psychology and computer science, providing detailed labels at both the session and comment levels. These labels capture essential aspects of cyberbullying, such as content type, purpose, directionality, and co-occurrence with other phenomena. The choice of this dataset aligns with our research objectives and enables an effective analysis of cyberbullying patterns on real-world social media platforms. An example of the comments that can be found in this dataset is presented in Table \ref{tab:dataset_example}\

\begin{comment}
\noindent
\textbf{Data Preprocessing.}
The dataset underwent several preprocessing steps to prepare it for analysis. Missing comment values in the original spreadsheet were replaced with empty strings and the comments were converted to string format. The target label indicating whether a comment is LGBTQ-related or not was derived from the original dataset and converted to integer format. To ensure consistent splits across experiments and robust evaluation, the preprocessed data was split into training and validation sets using stratified k-fold cross-validation with \( k=5 \) and a fixed random state of \( 42 \).
\end{comment}

\noindent
\textbf{Model Selection.}
We consider three pre-trained language models, namely Roberta \cite{liu2019roberta}, BERT \cite{devlin2019bert}, and GPT-2 \cite{RadfordGPT2}, to classify cyberbullying comments as either LGBTQ-related or Non-LGBTQ-related. For each model, we experimented with various configurations to evaluate their performance and robustness. These configurations include the original dataset without oversampling, oversampling using SMOTE (Synthetic Minority Over-sampling Technique) \cite{Chawla_2002}, and oversampling using ADASYN (Adaptive Synthetic Sampling) \cite{adasyn}. 
%Each of these configurations is evaluated using k-fold cross-validation to ensure the reliability of the results.

\noindent
\textbf{Model Training and Evaluation.}
For this study we considered the models, Roberta, BERT, and GPT-2, which were sourced from the Transformers library. We studied each model in an identical testing pipeline which includes preprocessing and tokenization using the model specific tokenizers, training the models to the desired number of epochs, and evaluating model performance based Accuracy, Precision, Recall, F1 score, and AUROC. For all the models, we utilized both SMOTE and ADASYN oversampling techniques with 5-fold cross-validation to address the potential class imbalance in the dataset and ensure the models perform consistently regardless of the training-validation split. The aforementioned oversampling techniques generate additional samples of the minority class to balance the class distribution and present the model with more opportunities to learn the underrepresented class. At the end of the training pipeline, each model has been tested across three experiment configurations, covering all possible combinations of the proposed oversampling techniques (original, SMOTE, and ADASYN).

\noindent
\textbf{Model Fine-tuning.}
We used a fine-tuning approach to enhance the performance of pre-trained language models for classifying LGBTQ+ cyberbullying comments. To use RoBERTa and BERT, these models were adapted with sequence classification heads. The models were then trained using the binary cross-entropy loss function and optimized with the AdamW optimizer. The learning rate was set to $5e-5$ and the models were trained for 3 epochs with a batch size of 16.
Adapting GPT-2 for classification required additional steps since it was originally designed for text generation. We utilized the GPT2ForSequenceClassification which adds a linear layer on top of the pooled output of GPT-2 enabling it to perform sequence classification. To handle variable-length sequences we used the EOS token as the padding token and set the maximum sequence length to 128 tokens.

\noindent
\textbf{Evaluation Metrics and Visualization.}
To assess the performance of the models, we used a range of evaluation metrics including accuracy, precision, recall, F1 score, and Area Under the Receiver Operating Characteristic curve (AUROC). These metrics provide a holistic view of the models' performance, considering different aspects such as correct classifications, false positives, and false negatives. In addition to quantitative metrics, we generated confusion matrices and visualization plots for each fold of the cross-validation. The performance metrics presented in the Experimental Results are the result of taking the average score across the 5 folds. 

\section{Experimental Results}

The experimental results of the three pre-trained language models; Roberta, BERT and GPT-2, for classifying cyberbullying comments as either LGBTQ-related or Non-LGBTQ-related are presented in Table \ref{tab:results}.

\begin{table}[h]
    \centering
    \begin{tabular}{|c|c|c|c|c|c|c|}
        \hline
        \multirow{2}{*}{MODEL} & \multirow{2}{*}{Oversampling} & \multicolumn{5}{c|}{Metrics} \\
        \cline{3-7}
        & & Accuracy & Precision & Recall & F1 & AUROC \\
        \hline
        \multirow{3}{*}{RoBERTa} 
        & Original & \textbf{0.9456} & 0.8712 & \textbf{0.6372} & \textbf{0.733} & 0.8592 \\
        & SMOTE & 0.938 & \textbf{0.8888} & 0.5664 & 0.6826 & 0.8376 \\
        & ADASYN & 0.9412 & 0.8658 & 0.6134 & 0.7044 & \textbf{0.8926} \\
        \hline
        \multirow{3}{*}{BERT} 
        & Original & 0.8922 & 0.4734 & 0.1092 & 0.172 & 0.7164 \\
        & SMOTE & 0.941 & 0.8792 & 0.5912 & 0.6942 & 0.8626 \\
        & ADASYN & 0.9298 & 0.8688 & 0.4786 & 0.611 & 0.871 \\
        \hline
        \multirow{3}{*}{GPT-2} 
        & Original & 0.9344 & 0.8556 & 0.5338 & 0.6506 & 0.8776 \\
        & SMOTE & 0.9032 & 0.7684 & 0.213 & 0.3218 & 0.7524 \\
        & ADASYN & 0.9114 & 0.8162 & 0.3332 & 0.4468 & 0.8078 \\
        \hline
    \end{tabular}
    \caption{Comparison of Different Models Across Various Metrics}
    \label{tab:results}
\end{table}

The RoBERTa model outperforms BERT and GPT-2 across most configurations, achieving the highest accuracy (0.9456) and F1 (0.733) scores without oversampling. The overall best AUROC score (0.8926) was also obtained with RoBERTa but using ADASYN oversampling. These results indicate RoBERTa's robustness in classifying LGBTQ-related cyberbullying comments. However, there are notable challenges and limitations in the models' performance, particularly in identifying cyberbullying instances, which we discuss below.

Despite the overall strong performance, the models, including RoBERTa, exhibit difficulties in accurately identifying cyberbullying instances, especially those targeted at LGBTQ+ individuals. Table \ref{tab:results_perclass} provides a per-class breakdown of precision, recall, F1 score, and AUROC for RoBERTa with different oversampling techniques. 

\begin{table}[h]
    \centering
    \begin{tabular}{|c|c|c|c|c|c|c|}
        \hline
        \multirow{2}{*}{Model} & \multirow{2}{*}{Oversampling} & \multirow{2}{*}{Class} & \multicolumn{4}{c|}{Metrics} \\
        \cline{4-7}
        & & & Precision & Recall & F1 & AUROC \\
        \hline
        \multirow{6}{*}{RoBERTa} 
        & \multirow{2}{*}{Original} & Non-LGBTQ+ & 0.952 & 0.986 & 0.968 & \multirow{2}{*}{0.8592} \\
        & & LGBTQ+ & 0.8712 & 0.6372 & 0.733 & \\
        \cline{2-7}
        & \multirow{2}{*}{SMOTE} & Non-LGBTQ+ & 0.938 & 0.994 & 0.962 & \multirow{2}{*}{0.8376} \\
        & & LGBTQ+ & 0.8888 & 0.5664 & 0.6826 & \\
        \cline{2-7}
        & \multirow{2}{*}{ADASYN} & Non-LGBTQ+ & 0.95 & 0.982 & 0.968 & \multirow{2}{*}{0.8926} \\
        & & LGBTQ+ & 0.8658 & 0.6134 & 0.7044 & \\
        \hline
    \end{tabular}
    \caption{Per-class Comparison of Roberta Models with Different Oversampling Techniques Across Various Metrics}
    \label{tab:results_perclass}
\end{table}

The per-class analysis highlights that although RoBERTa demonstrates strong performance in accurately identifying Non-LGBTQ+ bullying comments, as evidenced by the high precision and recall scores, it faces challenges when it comes to detecting LGBTQ+ bullying comments. This is reflected in the consistently lower recall scores for the bullying class across all experimental configurations. The model's difficulty in recognizing subtle and context-dependent instances of bullying leads to an increased number of false negatives, where LGBTQ+ bullying comments are incorrectly classified as Non-LGBTQ+ bullying. This limitation underscores the need for further research and development to enhance the model's ability to capture the nuances and complexities of LGBTQ+-related cyberbullying.

\begin{figure}[h]
    \centering
    \includegraphics[width=1\textwidth]{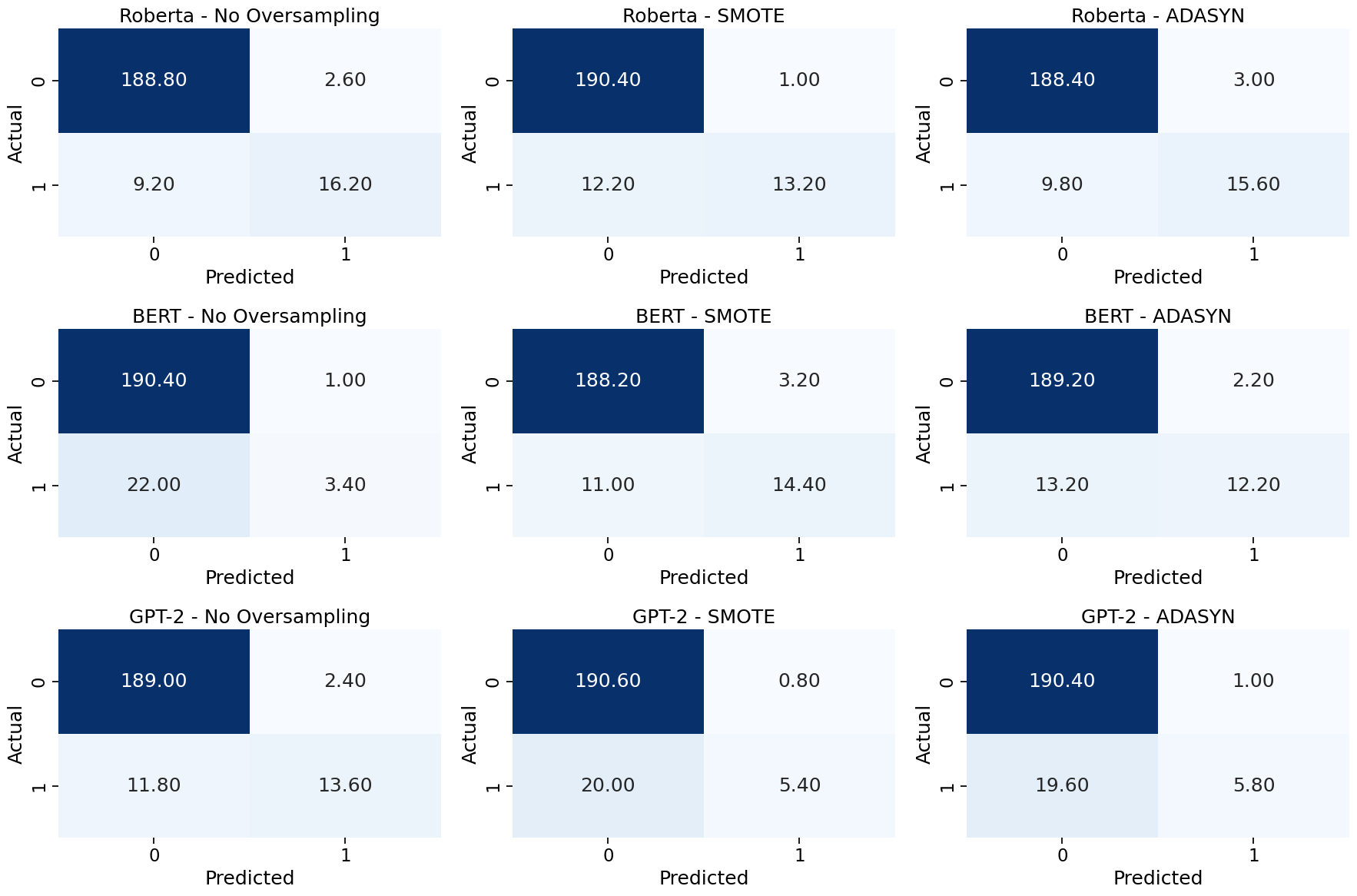}
    \caption{Above is a set of confusion matrices, one for each model and oversampling technique combination.}
    \label{fig:confusion_matrix}
\end{figure}

To further illustrate the models' performance, confusion matrices for each configuration are presented in Figure \ref{fig:confusion_matrix}. This figure includes the confusion matrices for each model and oversampling technique combination. The confusion matrices provide a comprehensive overview of the models' performance by displaying the actual and predicted values for both the Non-LGBTQ+ bullying (0) and LGBTQ+ bullying (1) classes. %The diagonal elements of the matrices represent the correctly classified instances, with true negatives (top-left) and true positives (bottom-right). The off-diagonal elements show the misclassified instances, with false positives (top-right) and false negatives (bottom-left).

Across all models and oversampling techniques, the true negative values (top-left) are consistently higher than the true positive values (bottom-right) indicating that the models perform better in identifying Non-LGBTQ+ bullying comments compared to LGBTQ+ bullying comments. The visualization also highlights the impact of oversampling techniques on the models' performance. For instance, SMOTE and ADASYN oversampling generally lead to a higher number of true positives compared to no oversampling, suggesting an improvement in the models' ability to detect LGBTQ+ bullying instances.

However, the confusion matrices also reveal potential areas for improvement, particularly in reducing the number of false negatives (LGBTQ+ bullying comments incorrectly classified as Non-LGBTQ+ bullying). This is evident from the relatively high values in the bottom-left cells of the matrices. Addressing this issue is crucial for developing more effective models that can accurately identify and mitigate cyberbullying targeting the LGBTQ+ community.

\section{Discussion}

\begin{table}
    \centering
    \begin{tabular}{l|l}
    \toprule
            \textbf{RoBERTA Mislabeled Comments} & \textbf{Outcome} \\
            \midrule
            No more calling me white trash or fag @p3bble & FP \\
            \hline
            @bacon I agree and @icrazyfan190 your an ass, \\
            gay people are still people and everyone should learn to respect \\
            them because we are all equal  & FP \\
            \hline
            @greenlady your what I like to call a top class bell end. You are a dyke! & FN \\
            \hline
            U like dick \#thatsurfuckinproblem & FN 
            \\
            \bottomrule
    \end{tabular}
    \caption{Samples of both False Positive (FP) and False Negative (FN) comments from RoBERTa.}
    \label{tab:false_negative_false_positive}
\end{table}

The experimental results presented in this study highlight the potential and limitations of state-of-the-art transformer models ( RoBERTa, BERT and GPT-2) for detecting LGBTQ+-related cyberbullying comments. While the models demonstrate overall effectiveness in distinguishing between Non-LGBTQ+ bullying and LGBTQ+ bullying comments, several technical aspects warrant further discussion.

\noindent
\textbf{Performance Evaluation and Challenges.}
RoBERTa consistently outperforms BERT and GPT-2 across various metrics, particularly in terms of accuracy and F1 scores. However, the confusion matrices reveal a disparity between true negatives (Non-LGBTQ+ bullying comments) and true positives (LGBTQ+ bullying comments), highlighting the models' tendency to favor Non-LGBTQ+ bullying classifications, leading to higher false negatives for LGBTQ-related Cyberbullying.
In Table \ref{tab:false_negative_false_positive}, we see can see two examples of misclassified false negatives comments. The first false negative might indicate that the model did not recognize the homophobic slur 'dyke', which is likely a symptom of insufficient training data. In the second example, the model likely struggled with the lack of context in the comment to confidently label it as LGBTQ+ bullying. The persistent challenge in accurately identifying LGBTQ+-related cyberbullying comments lies in the nuanced and context-dependent nature of such interactions. Cyberbullying targeting LGBTQ+ individuals often involves implicit language, sarcasm, and coded expressions, e.g., the false positives and false negative examples present in Table \ref{tab:false_negative_false_positive}, that are difficult for models to decipher, resulting in lower precision and recall rates for the bullying class.

\noindent
\textbf{Impact of Oversampling Techniques.}
The integration of oversampling techniques such as SMOTE and ADASYN shows a noticeable impact on the models' ability to detect bullying comments by providing more balanced training data. However, the issue of false negatives remains prevalent, suggesting that oversampling alone is insufficient to address the inherent complexities of LGBTQ+-related cyberbullying detection.

\noindent
\textbf{Future Work.}
Several techniques could be explored to further improve the performance of LGBTQ+ cyberbullying detection. These include: (1) integrating data features and model mechanisms that can capture deeper contextual and semantic nuances, (2) integrating multi-modal data including images, videos, and social network metrics (e.g., likes and shares) to enrich contextual data, (3) creating datasets that reflect a wider range of bullying scenarios, particularly those targeting LGBTQ+ individuals, and (4) exploring models that account for the sequence and timing of interactions to better capture intrinsic cyberbullying properties.

\noindent
\textbf{Limitations of the Current Study.}
The study's limitations include the dataset size and diversity and the single-platform focus on Instagram data. Addressing these limitations in future research is essential for developing more generalizable and effective cyberbullying detection systems.

\section{Conclusion}
This study evaluated RoBERTa, BERT, and GPT-2 for detecting LGBTQ+ cyberbullying on social media. RoBERTa outperformed BERT and GPT-2, but all models struggled to some extent with nuanced, context-dependent bullying instances. Oversampling techniques like SMOTE and ADASYN improved detection but did not fully resolve false negatives. Future work could focus on leveraging larger, more diverse datasets, incorporating multi-modal data, and developing fairness-aware training methodologies to enhance model effectiveness and create safer online spaces for the LGBTQ+ community.

\begin{credits}
\subsubsection{\ackname} This work was supported by NSF Awards \#2227488 and \#1719722 and a Google Award for Inclusion Research.
\end{credits}

\bibliographystyle{IEEEtran}
\bibliography{IEEEabrv, ref.bib}
\end{document}